\documentclass[runningheads]{llncs}

\usepackage[T1]{fontenc}

\usepackage{graphicx}
\usepackage{cite}
\usepackage{hyperref}
\usepackage[misc]{ifsym}

\newcommand{\subf}[2]{%
  {\small\begin{tabular}[t]{@{}c@{}}
  #1\\#2
  \end{tabular}}%
}

\begin{document}

\title{Grasping Partially Occluded Objects Using Autoencoder-Based Point Cloud Inpainting}

\titlerunning{Grasping Partially Occluded Objects Using AEPCI}

\author{Alexander Koebler(\Letter)\inst{1,5}\and
Ralf Gross\inst{5}\and
Florian Buettner\inst{1,2,3,4,5,}\textsuperscript{*}\and
Ingo Thon\inst{5}}

\authorrunning{A. Koebler et al.}

\toctitle{Grasping Partially Occluded Objects Using Autoencoder-Based Point Cloud Inpainting}
\tocauthor{Alexander Koebler, Ralf Gross, Florian Buettner, Ingo Thon}

\institute{
Goethe University Frankfurt, Department of Computer Science, Frankfurt, Germany \and
Goethe University Frankfurt, Department of Medicine, Frankfurt, Germany \and
German Cancer Consortium, Frankfurt, Germany \and
German Cancer Research Center Heidelberg, Heidelberg, Germany \and
Siemens AG, Germany\\
\email{\{alexander.koebler; ralf.gross; buettner.florian; ingo.thon\}@siemens.com}
}

\maketitle
\begin{abstract}
Flexible industrial production systems will play a central role in the future of manufacturing due to higher product individualization and customization. A key component in such systems is the robotic grasping of known or unknown objects in random positions. 
Real-world applications often come with challenges that might not be considered in grasping solutions tested in simulation or lab settings.
Partial occlusion of the target object is the most prominent. Examples of occlusion can be supporting structures in the camera's field of view, sensor imprecision, or parts occluding each other due to the production process.
In all these cases, the resulting lack of information leads to shortcomings in calculating grasping points.\newline
In this paper, we present an algorithm to reconstruct the missing information. 
Our inpainting solution facilitates the real-world utilization of robust object matching approaches for grasping point calculation.
We demonstrate the benefit of our solution by enabling an existing grasping system embedded in a real-world industrial application to handle occlusions in the input. With our solution, we drastically decrease the number of objects discarded by the process.

\end{abstract}

\section{Introduction}
\footnotetext{\textsuperscript{\rm * }Work done for Siemens AG}
Recent research efforts are paving the way towards robotic grasping of randomly positioned known \cite{li_finding} or unknown objects \cite{mahler_dex-net_2017, mahler_dex-net_2018}, with the potential to substantially increase the efficiency of many production processes.
However, key challenges inhibiting the utilization of robot grasping in real-world production seem to receive limited attention in research so far.
For example, in settings where not the entire target object is visible in the input scene, state-of-the-art approaches for unknown objects can only sample grasping points in sub-regions of the object captured by the vision system.
Conventional object matching approaches for known objects are even more affected by the lack of information and might not be able to determine any grasping point. Missing parts of the object in the input lead to a decline in the accuracy of the used surface or feature matching algorithms \cite{dantanarayana_object_2017}. Nevertheless, given complete input, conventional 3D object matching approaches have proven themselves in many real-world applications and offer a highly reliable calculation of grasping points. Their deterministic generation of the grasping points is often crucial for precise actions in pick and place applications. Thus, we aim to use machine learning (ML) to impute the missing information in the input such that robust conventional methods for grasping point detection can be used in more application areas.
Most often, the cause for missing information in the input data is the partial occlusion of the target object during the recording phase.
A simple remedy is re-scanning the input scene after removing the cause of the occlusion.
However, in many real-world applications, re-scanning is either undesirable as it usually causes an unproductive process prolongation or might even be impossible due to existing process constraints.\newline
In industrial applications, 3D models of the processed objects are commonly available as computer-aided design (CAD) models created in the own or the suppliers' product design process. In addition to their use in object matching approaches, these models can be utilized in combination with a simulation environment to generate a synthetic dataset.
In this work, we propose to use such generated data to train an auxiliary autoencoder to inpaint occluded areas of objects for robotic grasping. Our approach drastically reduces the impact on the downtime of the real-world assembly machine during the design and training phase.\newline
Our main contribution consists of the development and real-world demonstration of a novel algorithm to recover occluded object information in robotic grasping settings. This method enables the usage of established object matching algorithms for stable grasping point calculation without re-scanning the product or redesigning the production process.\newline
We make the following technical contributions:
\begin{itemize}
    \item We introduce a new algorithm to process unordered single-view point cloud data generated by an industrial laser scanner with 2D convolutional networks.
    \item We evaluate the use of style- and perceptual loss functions \cite{johnson_perceptual_2016, liu_image_2018} for autoencoder-based inpainting networks in the context of robotic grasping.
    \item We demonstrate how we bridge the gap between training the model on simulated data and deploying it in the real world.

\end{itemize}
\section{Related Work}
Our work combines the two research areas of robotic grasping and machine learning-based inpainting. To our knowledge, there are only a few approaches \cite{9095277} towards this combination.
Therefore, we will first introduce current methods for robotic grasping and discuss their robustness to occlusions in the input. Afterward, we will briefly introduce ML-based image inpainting and the issues associated with transferring those methods to point cloud data.
\paragraph{Robustness of Grasping Algorithms to Occlusions:}
Data-driven methods to determine grasping points vary widely in terms of their build-in robustness to missing information in the input scene.\newline
Known approaches can be distinguished by the required knowledge about the objects that should be grasped \cite{bohg_data-driven_2014}.
If the exact object shape is known, common solutions rely on manually labeling optimal grasping points for specific objects. This can, for example, be done with the help of CAD models. The 3D models of the objects and thereby the predetermined grasping points can then be matched to input point clouds by various conventional algorithms or a combination of them \cite{dantanarayana_object_2017}. Examples include relying on a subset of the points as features such as RANSAC \cite{fischler_random_1981} or the entirety of the point cloud such as ICP \cite{schoenemann_generalized_1966}. These matching algorithms are heavily affected by missing information in the input point cloud with respect to the object of interest since this can cause a lack of matching key features between the recorded point cloud and the available 3D models. This issue has motivated research efforts towards increasing the robustness of those methods to occlusions and cluttered environments \cite{papazov_efficient_2010, papazov_rigid_2012} as well as the development of more recent ML-based pose estimation methods \cite{8886003}.\newline
Recent approaches for handling unknown objects select grasping point candidates on the visible areas based on local attributes and subsequently classify the most promising grasping point by a trained ML model \cite{mahler_dex-net_2017, mahler_dex-net_2018}. Although these methods might be more robust in the case of cluttered and occluded scenes, they lack the robustness of conventional matching algorithms in non-occluded cases. For example, they do not generate deterministic grasping points for known objects. Those, however, are required for subsequent processing steps, such as the precise placing of known objects.\\
\paragraph{Inpainting Approaches for Point Clouds:}
Our solution aims to enhance existing object matching solutions by an auxiliary autoencoder-based inpainting step. This preprocessing step ensures that an object's complete shape is present in the point cloud.\newline
The reconstruction of missing information in natural images or videos is a well-researched task relying on ML- \cite{vo_2018,liu_image_2018} and non-learning-based inpainting \cite{criminisi_region_2004}.\newline
Per-pixel reconstruction loss functions such as mean-squared-error (MSE) or mean-average-error (MAE) in the ML-based inpainting methods often do not satisfy the complicated semantic structure of the images. Thus, \cite{johnson_perceptual_2016} utilizes a loss network trained on a supervised classification task to evaluate high-level abstractions of the reconstructed image. By using autoencoder architectures for super-resolution and style-transfer \cite{johnson_perceptual_2016} as well as for inpainting \cite{liu_image_2018, vo_2018} of natural images, these loss terms resulted in significant improvements in the removal of image artifacts and in the visual perception of the results. Other approaches \cite{iizuka_globally_2017,yu_free-form_2019} use additional adversarial loss functions, which are inspired by generative adversarial networks (GANs) \cite{goodfellow_generative_2014}, to include higher-level abstractions in the reconstruction objective.\newline
Using generative models for unordered 3D point clouds instead of images is a less researched field \cite{xie_generative_2021,mandikal_dense_2019}. The non-equidistant grid and irregular format of the recorded point clouds do not allow for directly applying convolutional neural networks (CNNs) \cite{qi2017pointnet}, which are often used in vision use-cases. The authors in \cite{Zhao2020PUINetAP} are concerned with reconstructing small holes in point clouds rather caused by noise than by occlusions. This does not satisfy our requirement of reconstructing significant coherent portions of more than $80\%$ of the object point cloud.

\section{Problem Description}
\begin{figure}[tb!]
\centering
\includegraphics[scale=0.7]{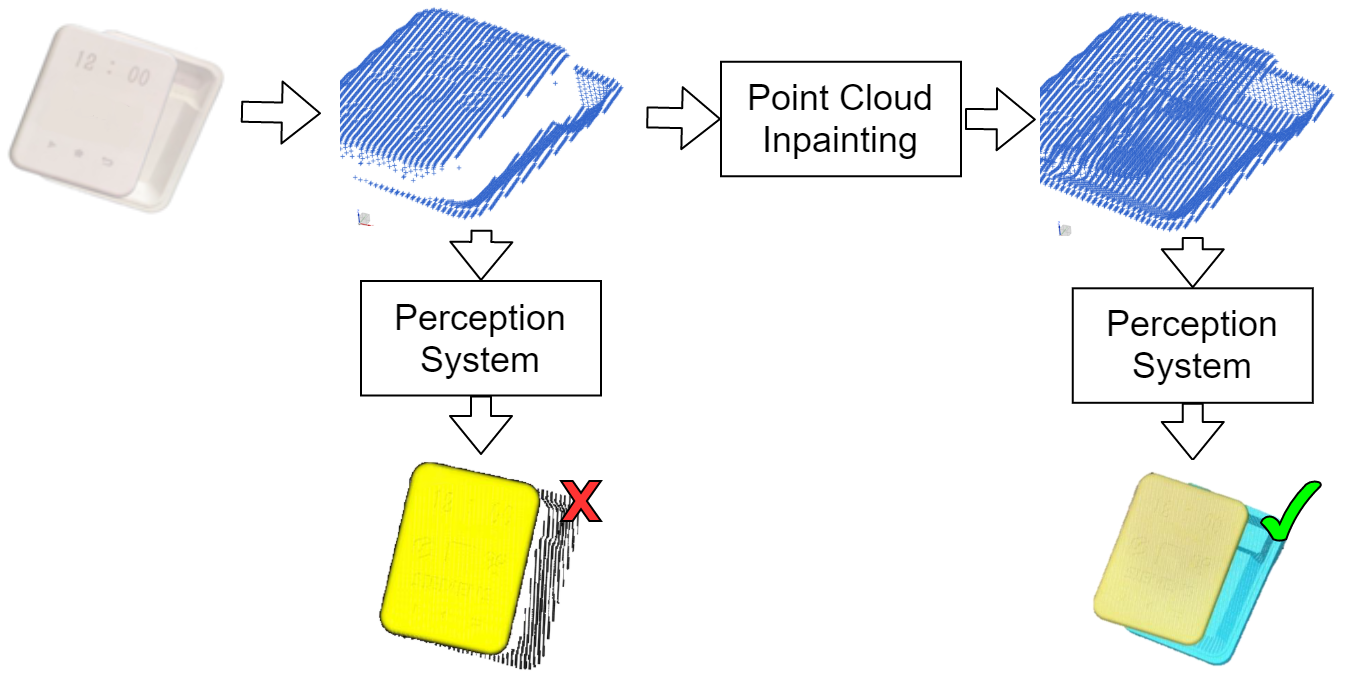}
\caption{Investigated problem setting for point cloud inpainting. As shown on the left, the laser scanner can only capture a small portion of the geometry of the occluded object. This is insufficient for the 3D matching algorithm that calculates the grasping points. Our inpainting solution outputs the complete point cloud on the right with the reconstructed lower object. With that, the surface-based matching algorithm can estimate the pose and orientation of both objects and determine corresponding grasping points.} \label{problem}
\end{figure}

In the context of a real-world assembly machine that includes a grasping application, we face the following problem: During the process, pairs of associated objects drop onto a conveyor belt. The objects have a size of about $30$mm $\times$ $25$mm and form the front and back parts of the assembled product. The objects are transported towards a robot arm and should subsequently be grasped with a suction gripper. For this purpose, a laser scanner prior to the robot arm scans the objects on the conveyor belt as they pass through the beam. Re-scanning the scene after the execution of a grasping process is not viable. It would require a time-consuming retraction of the conveyor belt to pass the object through the laser beam repeatably. The laser scanner generates unordered point clouds from a single point of view.
The geometry of the objects is known, and a system capable of determining appropriate grasping points on the objects is in place. In the following, this system will be referred to as the perception system. The perception system uses a surface-based 3D matching algorithm and predefined deterministic grasping points for the known objects. The system works reliably if the laser scanner captures the entire geometry of the target objects.\newline
When the objects drop on the conveyor belt, they can end up laying upon each other. Thus in some cases, one object partly occludes the other one. The resulting degree of occlusion is random and varies widely. The perception system can not determine grasping points for a large majority of occurring occlusions.\newline
Our objective, shown in Figure \ref{problem}, is to develop a solution to reconstruct the missing parts of the lower object after the scanning process.
The reconstruction quality of object parts must be sufficient for the subsequent surface-matching algorithm to match the object with acceptable accuracy.
The perception system requires well-restored key features of the objects, such as edges and corners. These are often not sufficiently captured by common reconstruction metrics such as the MSE or the MAE. For this reason, we evaluate the reconstruction quality in a task-specific manner. Our final evaluation criterion for the overall system is the rate of successful grasps in the case of occlusions. In addition, a visual inspection of the reconstruction can offer an estimation of its quality and the later gasping success rate.\newline
The event that objects occlude each other in the input scene is frequent enough that it causes a significant amount of discarded objects. However, gathering a huge dataset for training a reconstruction model would require a long-term operation of the production machine in a sub-optimal state. Therefore, the amount of available real-world data is limited.
Furthermore, collecting the point cloud of the non-occluded lower object in the same pose and orientation in the regular process is not possible. This is because the scanning of the input scene and the grasping of the robot are locally separated. Moreover, after the robot grasped the top object, re-scanning the scene with the no more occluded lower object would not yield the ground truth for the previously occluded object but a scan of the now isolated object after the robot operation or the conveyor retraction potentially impacted its positioning.\newline

\section{Methodology}
We propose using a processing pipeline including an autoencoder-based inpainting step to reconstruct missing information in partially occluded point clouds. In this section, we will elaborate on the components of the pipeline and substantiate our design decisions.\newline
The scanned scene's height with two objects is less than $15$mm, even if the objects fall on top of each other. This results in only small deviations in the height-dependent spacing of the grid generated by the scanner. Thus, we consider the information loss by interpolating the point clouds to an equidistant grid in the $(x,y)$-coordinates as negligible. The representation of the object on an equidistant grid allows for using regular image processing approaches such as CNNs. Thus, we can circumvent the use of more complicated methods that can handle unordered point clouds on non-equidistant grids \cite{qi2017pointnet}.
In our case, the considered objects are relatively simple, with only a few defining structures that must be preserved. For this reason, we decided on an autoencoder U-Net architecture \cite{ronneberger_u-net_2015} for the image inpainting process.\newline 
\subsection{Dataset Generation}
\begin{figure}[tb!]
\centering
\begin{tabular}{c c c c}
\subf{\includegraphics[height=0.15\textheight]{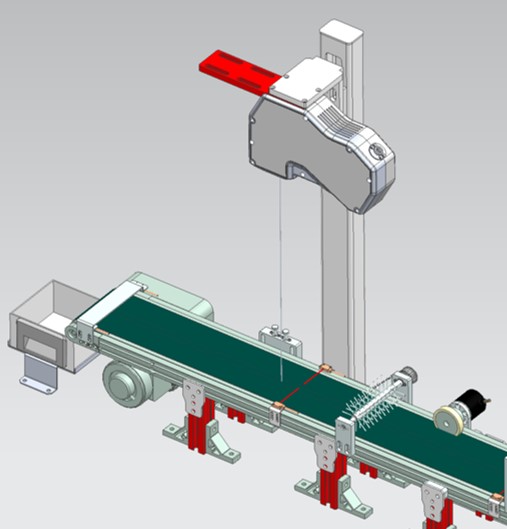}}
     {(a)}
&
\subf{\includegraphics[height=0.15\textheight]{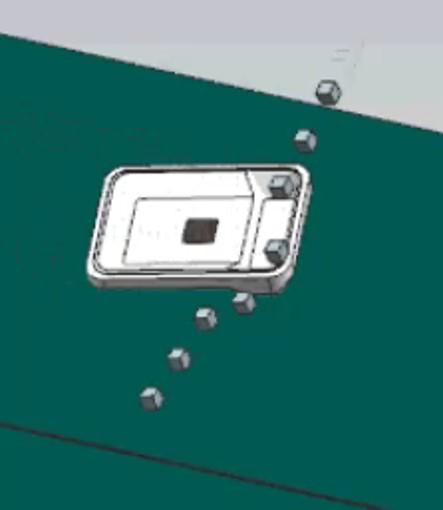}}
     {(b)}
&
\subf{\includegraphics[height=0.15\textheight]{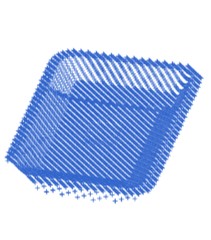}}
     {(c)}
\end{tabular}
\caption{Simulation environment for generating a synthetic training dataset of occluded scenes. In (a), the green conveyor belt and the laser scanner are shown. (b) depicts the scanning process of a single object, where the small grey cubes illustrate the beams of the laser scanner. The resulting synthetic point cloud is shown in (c).}\label{simulation}
\end{figure}

A method was developed to generate and record realistic instances of object occlusions via simulation. This synthetically generated data allows us to train the autoencoder model without the need to record a ground truth point cloud of the complete lower object in the real process.
The considered machine was already available as a digital twin in a product lifecycle management software with an integrated simulation toolkit. The relevant components shown in Figure \ref{simulation}, consisting of the conveyor belt, the laser scanner, and the product objects, could be utilized to generate the training dataset.
During the simulation, the objects are placed upon each other in random relative positions within a predefined design space. The integrated physics simulation subsequently causes objects to fall into physically valid positions. This allows for realistic occlusion patterns. The dataset is generated by recording the scene where both objects are present, and parts of the lower object are hidden from the scanner's view. Afterward, the top object is removed and the scene only consisting of the lower object is re-scanned, generating a ground truth point cloud of the complete lower object in the corresponding pose. Furthermore, an assignment between points and objects is generated for the point cloud with both objects.
\subsection{Data Processing Pipeline}
\begin{figure}[tb!]
\centering
\includegraphics[scale=0.06]{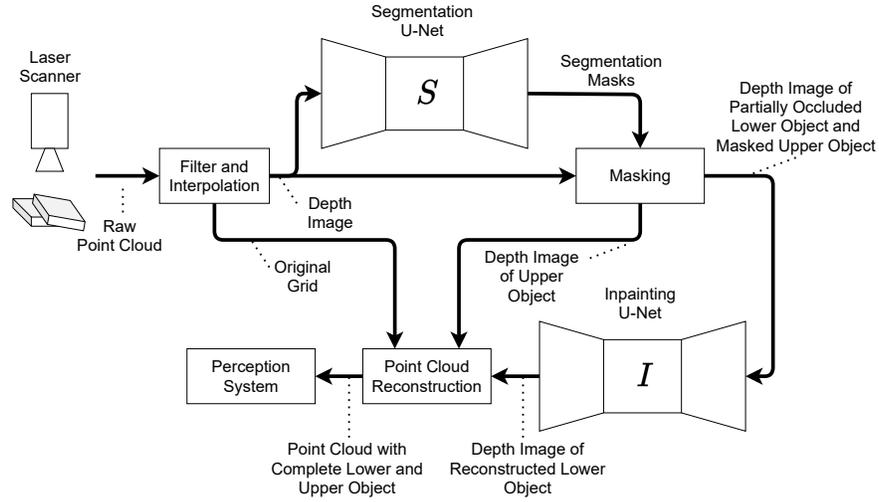}
\caption{Data processing pipeline with resulting in- and outputs of the deployed point cloud inpainting solution} \label{processing_pipeline}
\end{figure}

\begin{figure}[tb!]
\centering
\begin{tabular}{c c c}
\subf{\includegraphics[width=0.3\textwidth]{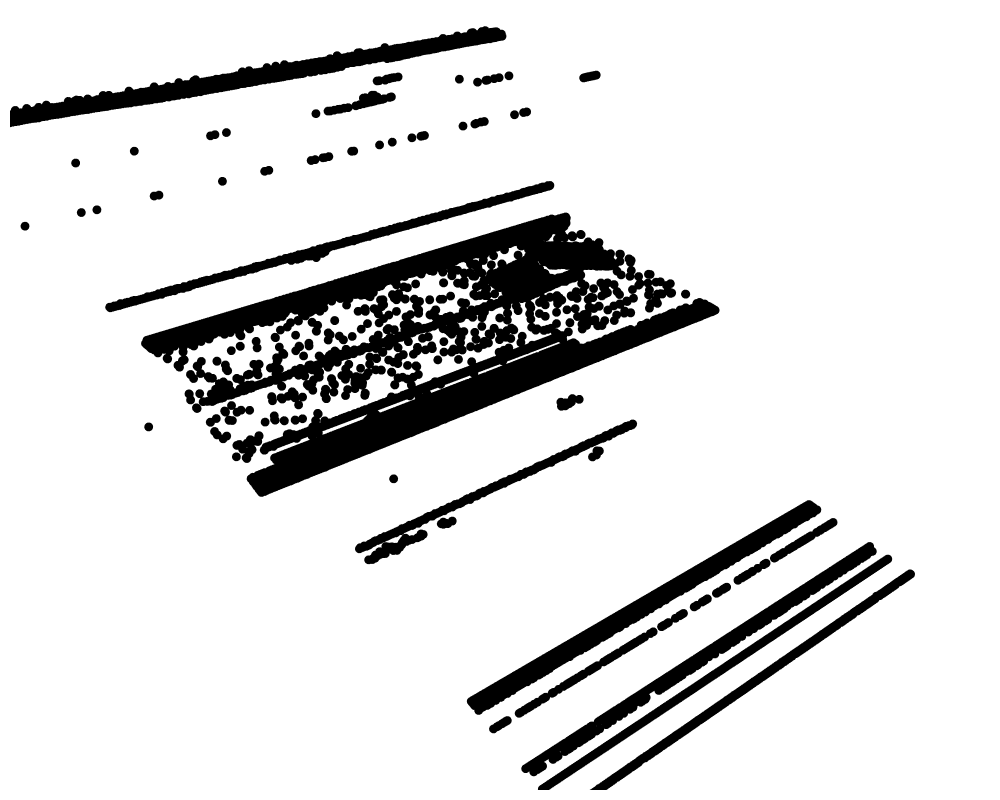}}
     {(a)}
&
\subf{\includegraphics[width=0.3\textwidth]{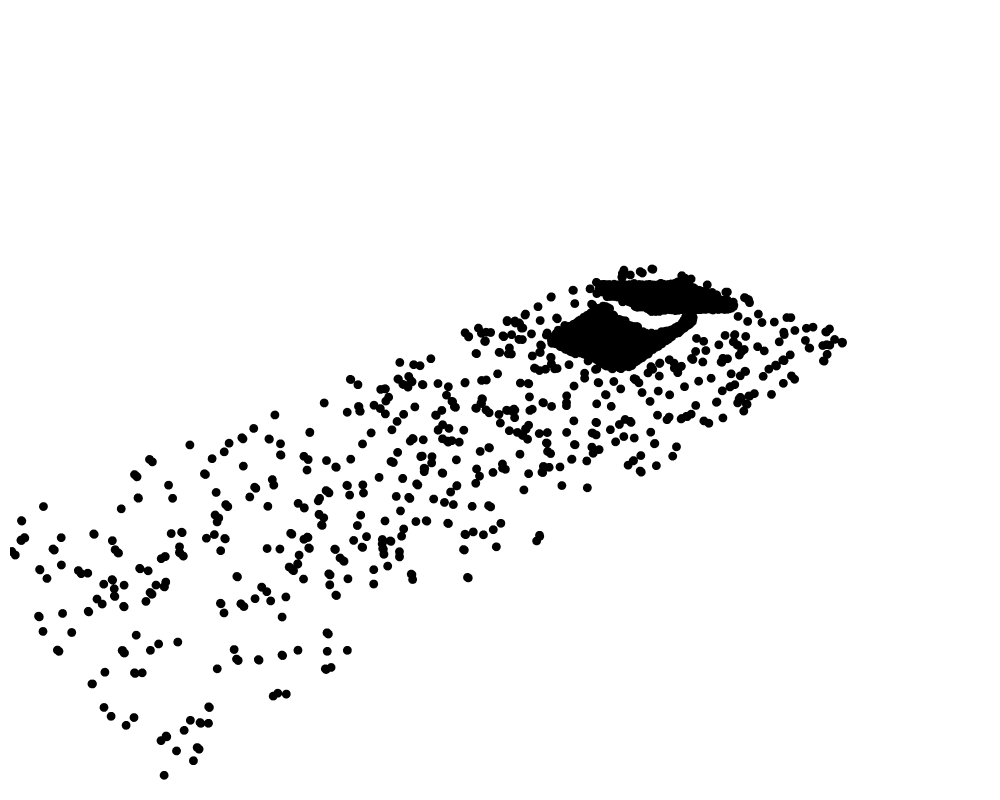}}
     {(b)}
&
\subf{\includegraphics[width=0.3\textwidth]{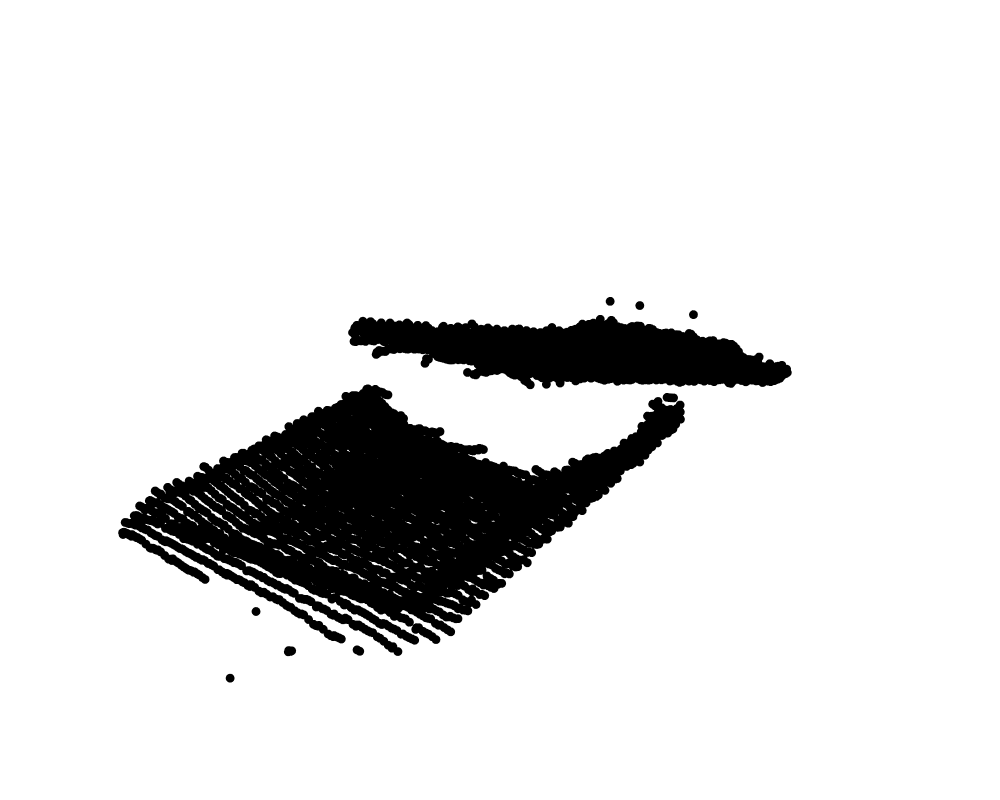}}
     {(c)}
\end{tabular}

\caption{Filtering steps for the real point clouds. The laser scanner captures different forms of noise and distractions in the raw input point cloud (a), such as the floor on the lower right and the wall on the left side. After cropping the distractions from the recorded view, sparse noise on the conveyor belt is left (b). By removing the sparse noise, only the dense clusters of the objects themselves remain (c).}\label{raw_input}
\end{figure}

We aimed to simplify the inpainting task and thereby reduce the performance gap we have to cross when we deploy our solution solely trained in simulation to the real-world process. Hence, we split the overall task into multiple simpler sub-tasks, which results in the processing pipeline depicted in Figure \ref{processing_pipeline}.\newline
The in Figure \ref{raw_input} depicted point clouds recorded by the real laser scanner suffer from two kinds of noise.
We apply predefined cropping windows to remove dense clusters from the point cloud that are caused by the conveyor rails, the floor, or the wall. The remaining noise after this step appears sparsely spread. Therefore, a clustering algorithm is used to remove points that do not correspond to the dense clusters formed by the objects. We use the DBSCAN algorithm \cite{ester_density-based_1996} for this purpose.
The required parameters for the filtering operations are tuned manually on a small amount of real input point clouds. The cropping boundaries in the $x$-direction orthogonal to the conveyor are set to a frame of $92$mm to remove the conveyor rails. Furthermore, only a window of $18$mm in depth direction $z$ is considered to remove artifacts caused by the wall and the floor. The maximal distance of neighboring points for the DBSCAN algorithm is set to $4$mm. Since the top objects often protrude at an angle, the two objects do not always form one coherent cluster. Found sub-clusters are considered to correspond to one of the objects if they include at least $500$ points. The remaining complete object cluster must consist of at least $6000$ points after the filtering step to detect errors in the scanning process.\newline
In a consecutive step, the resulting point cloud is interpolated to an equidistant grid. The generated grid has a resolution of $1.1$mm in $y$-direction orthogonal to the laser beam and $0.3$mm in the $x$-direction parallel to the beam. These distances are independent of the $z$-value at a specific point. They represent the original grid close to the height of a flat non-occluded object. Considering the height and width of two objects in all occurring configurations, a grid with $256$ points in $x$-direction and $64$ points $y$-direction is generated.
For the interpolation step, the points in the unordered original point cloud are matched by their $(x,y)$-coordinates with their nearest-neighbor in the generated equidistant grid by using a k-d-tree algorithm \cite{10.1145/361002.361007}.
The generated pairs between the $(x, y)$-values of the original and the equidistant grid are subsequently used to interpolate the $z$-values for the point cloud on the equidistant grid.
The resulting point clouds are interpreted as depth images, and only the $z$-values are used for further processing steps.\newline
We include a segmentation network to extract the top object for later reconstructing the entire input scene. Furthermore, using the segmentation network's output, the pixels corresponding to the top object in the original image are set to a constant value of $15$mm above the height of a single object. We expect this step to support the inpainting network to distinguish between the image areas that are relevant for the inpainting and to identify the structures corresponding to the lower object.\newline
After the inpainting model outputs the depth image of the reconstructed lower object, a post-processing step is necessary to recombine the entire point cloud.\newline
The pixels in the output depth image that correspond to the occluded object and not to the background or interpolation artifacts are identified by a dynamic thresholding algorithm \cite{otsu_threshold_1979}.
The top object segmented from the original depth image and the reconstructed lower object are subsequently mapped back to the original non-equidistant grid. For this purpose, the pixel-to-position mapping generated by the k-d-tree in the interpolation step is reused. The resulting point cloud can further on be processed by the perception system.\newline
\subsection{Segmentation}
For the segmentation task, a commonly used approach is utilized and implemented based on \cite{yakubovskiy_segmentation_2019}. For the segmentation network we decided for a U-Net architecture with an EfficientNetB2-backbone model \cite{tan_efficientnet_2020}. For training, we only use synthetic point clouds. The depth image of the scene where parts of the lower object are occluded is provided as input. The labels are the simulated point-to-object assignments that are also mapped to the generated equidistant grid. This results in 2D segmentation maps with the three categories: top object, occluded object, and background.
\subsection{Inpainting}
\begin{figure}[tb!]
\centering
\includegraphics[scale=0.06]{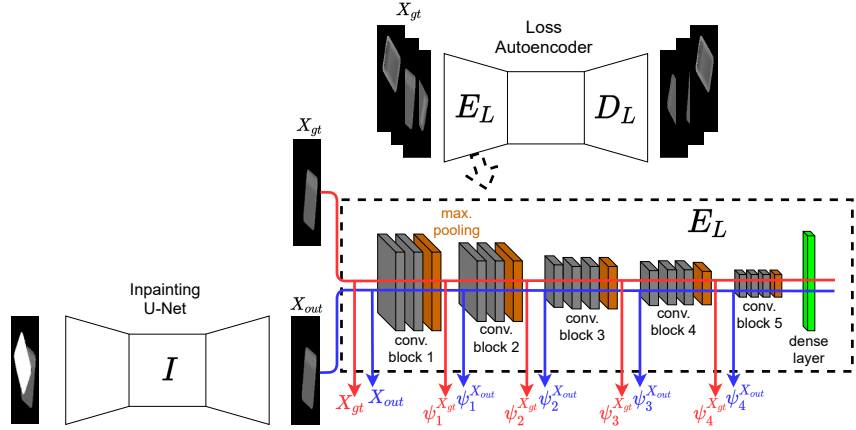}
\caption{Training architecture for the inpainting model using a VGG16 loss-network. The loss-network $E_L$ is trained unsupervised by learning equivalence on the depth image of the complete lower object. The inpainting model $I$ is trained to reconstruct the complete lower object from the depth image, including the masked top object and the partially occluded lower object. The intermediate representations of the ground truth image $\psi_p^{X_{gt}}$ and the reconstructed image $\psi_p^{X_{out}}$ for the style and perceptual loss are taken after the first four convolutional blocks.} \label{loss_fig}
\end{figure}

We also used a U-Net architecture with a VGG16-like \cite{simonyan_very_2015} backbone for the inpainting model.
Especially if significant parts of key features such as edges or corners are occluded and missing in the reconstruction, we assume that simple per-pixel loss functions such as MSE do not capture those differences sufficiently. However, these features are essential for the perception system's 3D matching algorithm. Thus, we have evaluated the style and perceptual loss functions applied for super-resolution and style-transfer in \cite{johnson_perceptual_2016} and image inpainting in \cite{liu_image_2018}.\newline
The proposed loss function relies on high-level intermediate representations extracted from a pretrained loss-network in addition to a per-pixel difference in the input space.
In contrast to \cite{liu_image_2018,johnson_perceptual_2016} we do not apply these losses to natural images similar to common public datasets. Thus, for our loss-network, we can not use a model that is trained on a supervised classification task on a dataset such as ImageNet.\newline
For this reason, we train our VGG16 loss-network as an autoencoder in an unsupervised manner. The autoencoder is trained such that the input and the expected output are the synthetic depth images of the ground truth lower object. The unsupervised task does not necessarily result in relevant features for a classification task. However, we expect the bottleneck to capture coherent high-level features of the object, such as edges and corners. Those features should then be represented by the intermediate representations generated by the network.\newline
As shown in Figure \ref{loss_fig}, the representations $\psi_p^{X}$ used in the style and perceptual loss are the output of the max-pooling layers. In particular, the first four convolutional blocks $p$ of the VGG16 loss-network are used.
The perceptual loss is the sum of the norm of the difference of the first four intermediate representations for the ground truth image $X_{gt}$ and the by the inpainting network generated image $X_{out}$:
\begin{equation}
\mathcal{L}_{perceptual} = \sum_{p = 1}^{P} ||\psi_p^{X_{out}} - \psi_p^{X_{gt}}||_1 \cdot \frac{1}{C_pH_pW_p}
\end{equation}
The feature maps $\psi_p^{X}$ are of size $C_p \times H_p \times W_p$.\newline
For the style loss, the auto-correlation matrix of the representations is calculated first. The norm of the resulting gram matrices of size $C_p \times C_p$ is normalized by the size of the feature maps
\begin{equation}
\mathcal{L}_{style} = \sum_{p = 1}^{P} \frac{1}{C_pC_p} \cdot ||\frac{1}{C_pH_pW_p} \cdot ((\psi_p^{X_{out}})^T \cdot (\psi_p^{X_{out}}) - (\psi_p^{X_{gt}})^T \cdot (\psi_p^{X_{gt}}))||_1 \, .
\end{equation}
During our experiments we considered this loss with and without the normalization factor of $\frac{1}{C_pC_p}$.
As in \cite{liu_image_2018} and different to \cite{johnson_perceptual_2016} we have used the $L_1$-norm in the style and perceptual loss terms and the per-pixel loss component
\begin{equation}
\mathcal{L}_{pixel} = \frac{1}{CHW} \cdot ||X_{out} - X_{gt}||_1 \, .
\end{equation}
If we normalize the perceptual loss by the size of the feature maps and the style loss by the size of the gram-matrix, we also have to normalize the per-pixel loss term by the size of the image in input space $C\times H\times W$ to keep all terms in the same order of magnitude.
The combined perceptual and style-based loss (PSBL) results as:
\begin{equation}
\mathcal{L}_{psbl} = \mathcal{L}_{pixel} + \alpha \mathcal{L}_{perceptual} + \beta \mathcal{L}_{style}
\end{equation}
The weight of the perceptual and style loss is determined by the hyperparameters $\alpha$ and $\beta$, respectively.
\section{Experiments}
In the experiments section, we want to focus on the performance of the inpainting network before we end with a performance analysis of the entire pipeline on the real machine.
\subsection{Training Procedure}
We first pretrain all models with MSE loss for $200$ epochs to establish a proper initialization of the network weights. Subsequently, we fine-tune the models for the same amount of epochs on the evaluated loss functions. We use an Adam optimizer and a learning rate of $0.0001$ for both training phases.\newline
The loss-network is trained for $300$ epochs with MSE loss. The bottleneck dimension for the autoencoder-based unsupervised pretraining is set to $16$. This value is empirically set so that the autoencoder can still learn sufficient equality but must learn high-level abstractions to handle the narrow bottleneck at the same time.\newline
The generated synthetic dataset consists of $21\,000$ examples. It is split into $16\,800$ training samples and $4\,200$ test samples.
Of the $16\,800$ training samples $3\,360$ are used as a validation set to find the best checkpoint based on the peak signal-to-noise ratio (PSNR) during training. This results in an actual training dataset of $13\,440$ samples.
\subsection{Sensitivity to Initialization}
We receive very unstable results for training the models from scratch with random initialization. In case of a training collapse, the model's outputs converge within the first ten epochs to an empty image which can easily be detected programmatically. For the different losses, this happens for a different subset of random seeds. When we evaluate using the same $25$ random seeds, the models trained with MAE loss collapse in $80\%$ of the time, with MSE loss in $50\%$, with PSBL loss without normalization in $60\%$, and with layer-wise loss normalization in $80\%$, we assume this is due to the high ratio of empty background in most images.
If the models are initialized by pretraining on MSE, they do not collapse using other losses during the fine-tuning step.\newline
\subsection{Performance on Synthetic Data}
\begin{figure}[tb!]
\centering
\begin{tabular}{c| c c c c}
\subf{\includegraphics[width=0.18\textwidth]{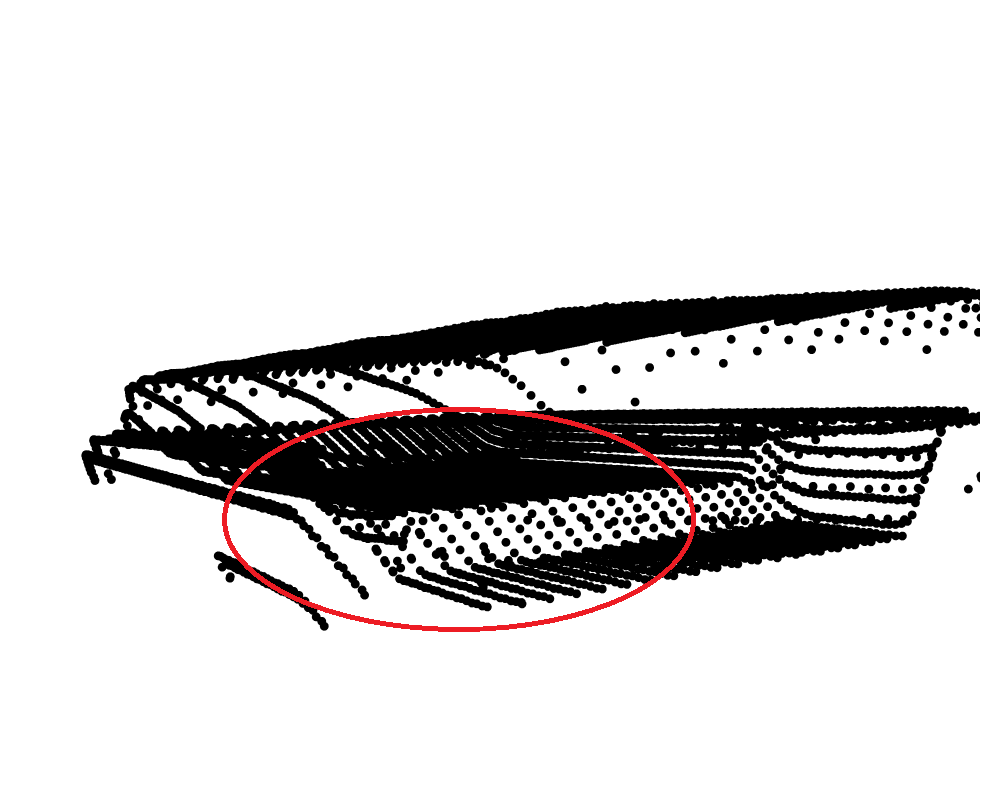}}
     {}
&
\subf{\includegraphics[width=0.18\textwidth]{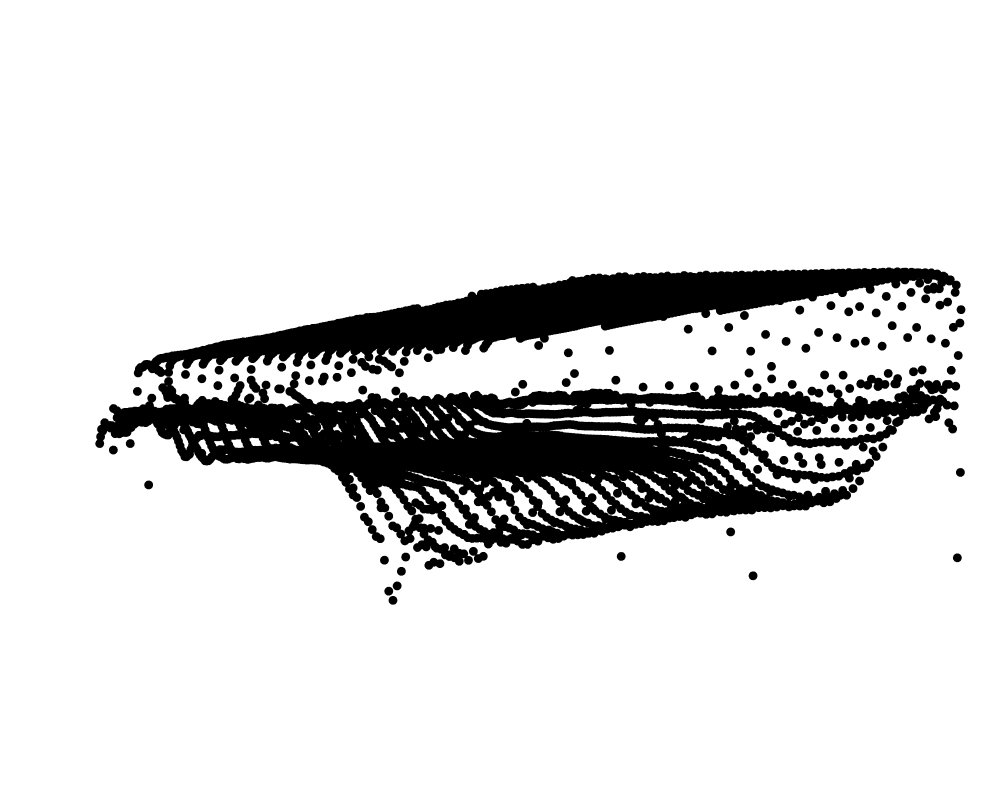}}
     {}
&
\subf{\includegraphics[width=0.18\textwidth]{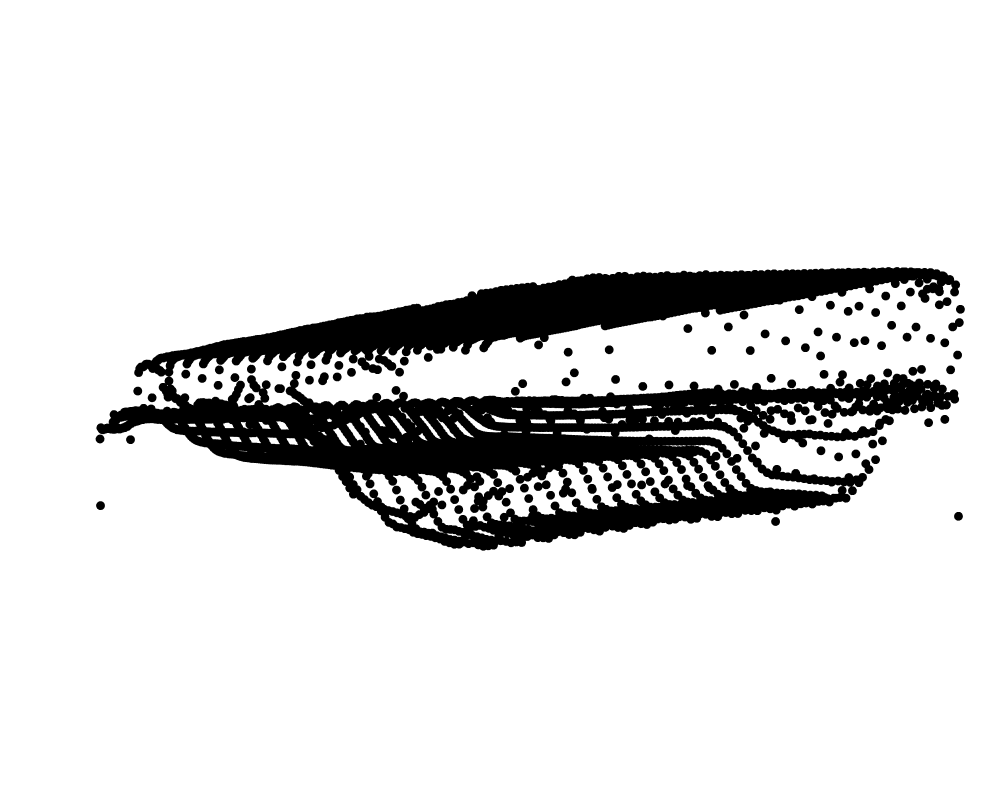}}
     {}
&
\subf{\includegraphics[width=0.18\textwidth]{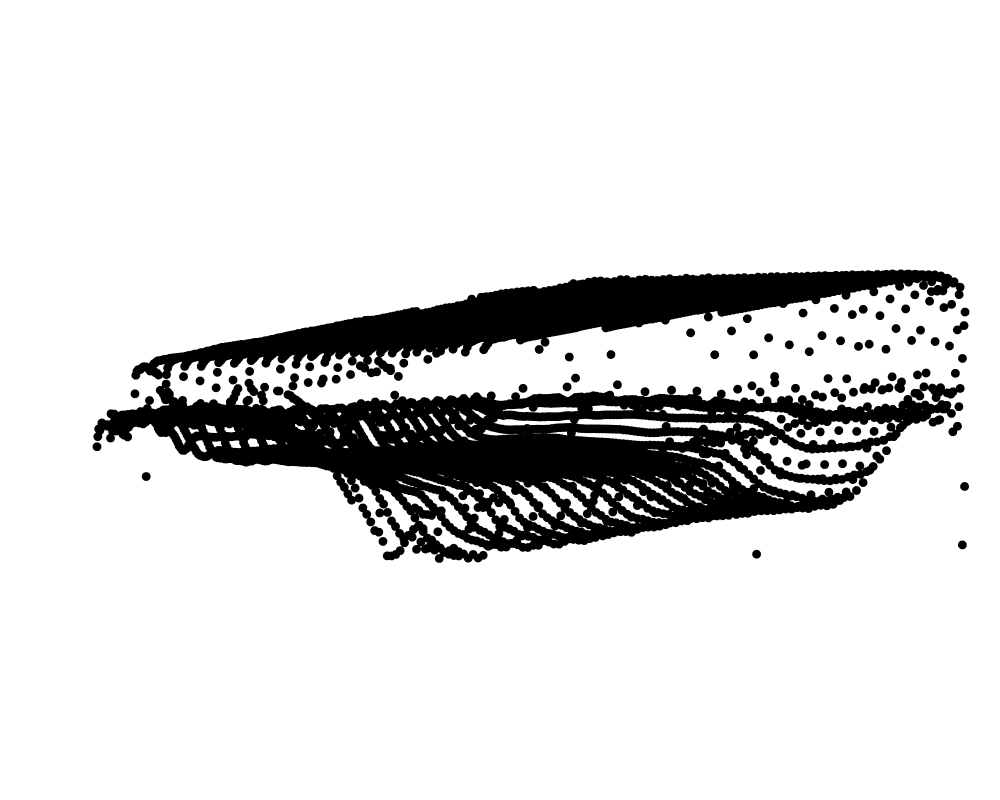}}
     {}
&
\subf{\includegraphics[width=0.18\textwidth]{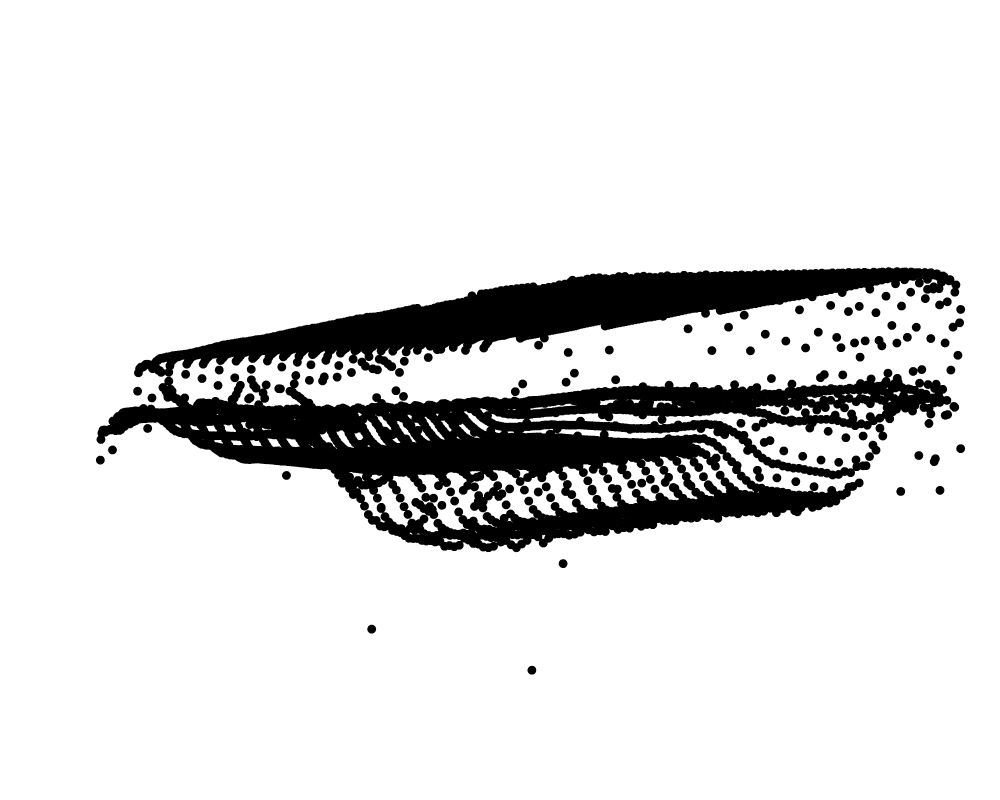}}
     {}
\\
\subf{\includegraphics[width=0.18\textwidth]{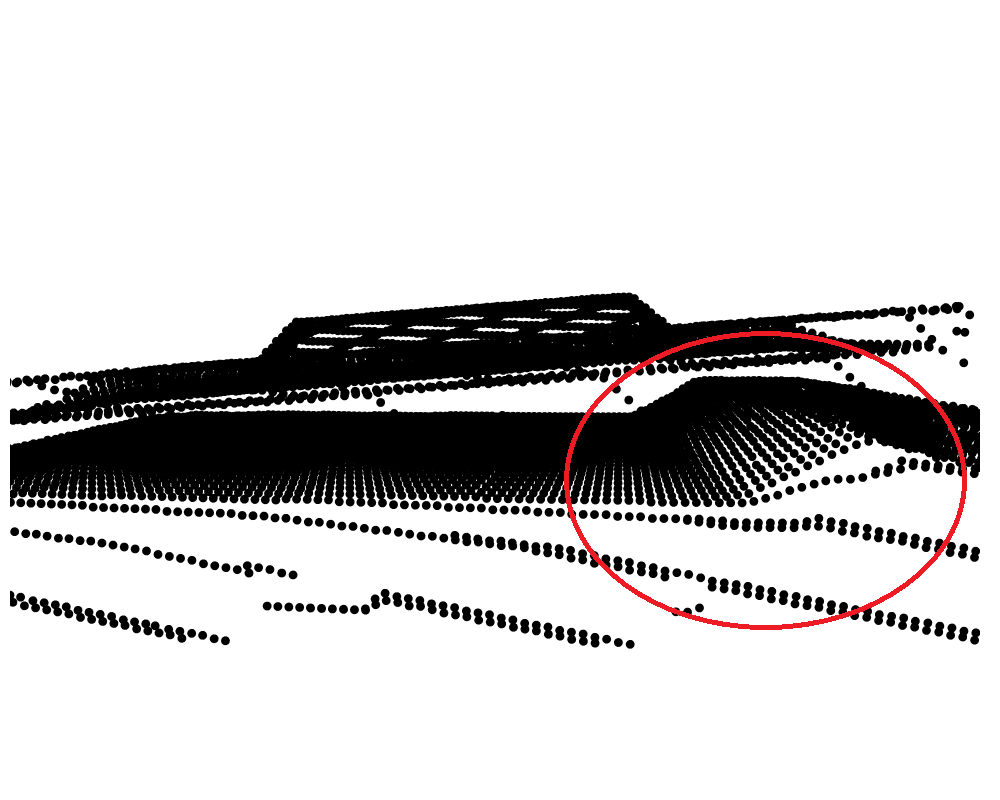}}
     {ground\\ truth}
&
\subf{\includegraphics[width=0.18\textwidth]{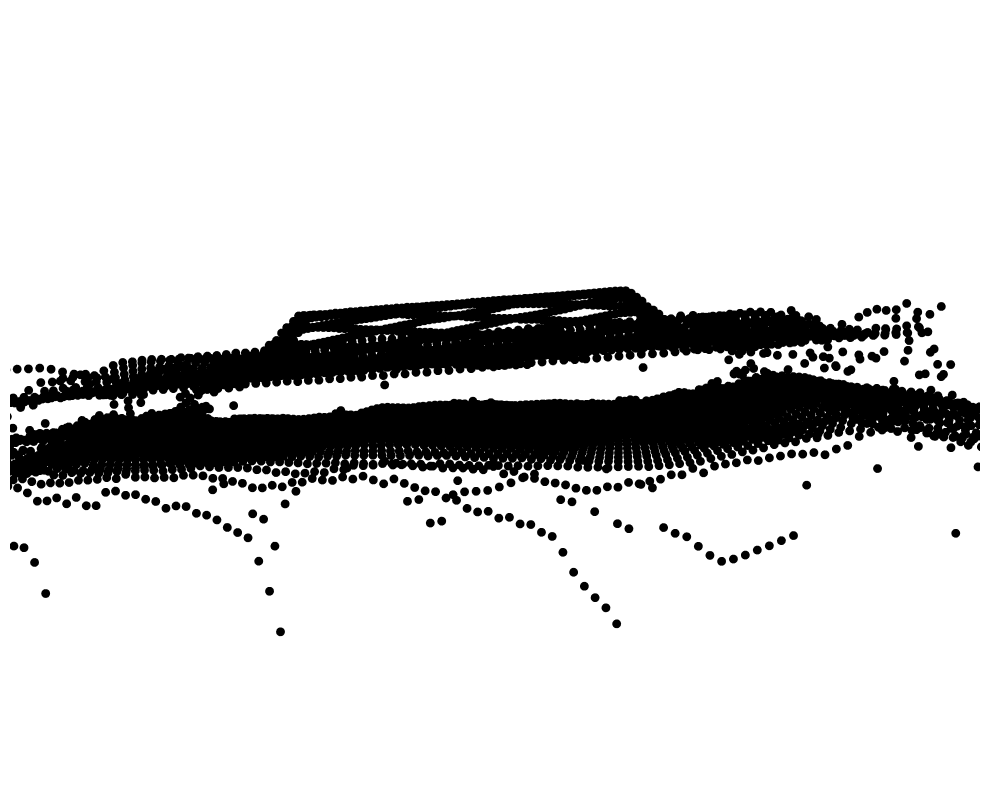}}
     {MSE}
&
\subf{\includegraphics[width=0.18\textwidth]{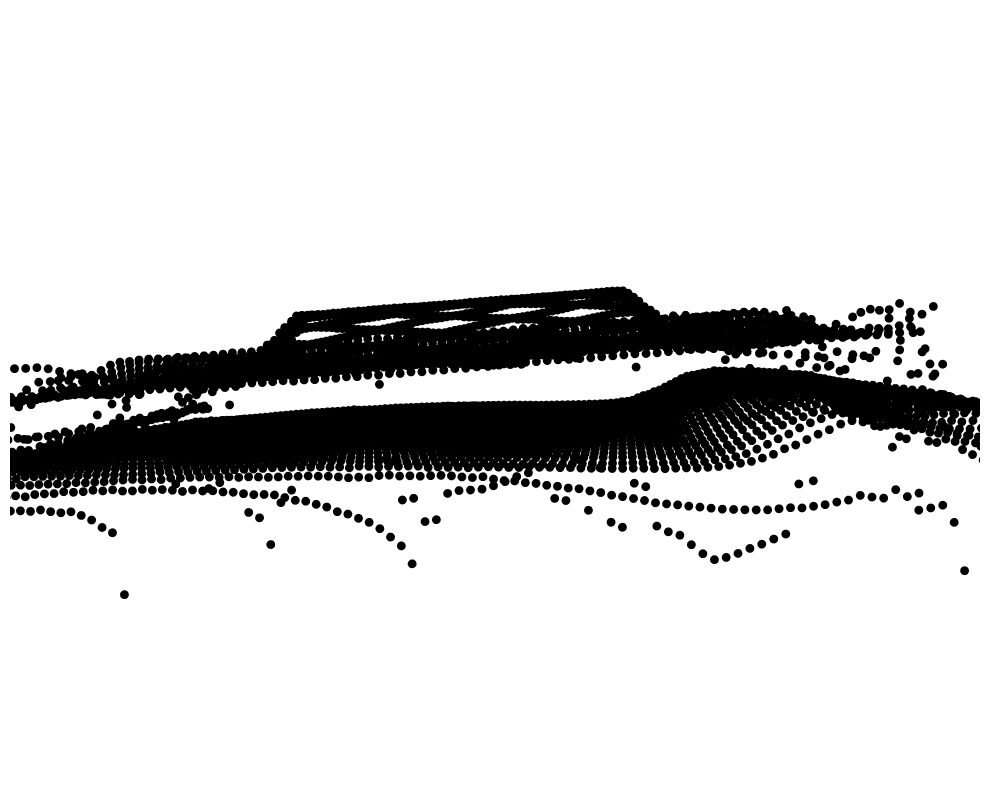}}
     {MAE}
&
\subf{\includegraphics[width=0.18\textwidth]{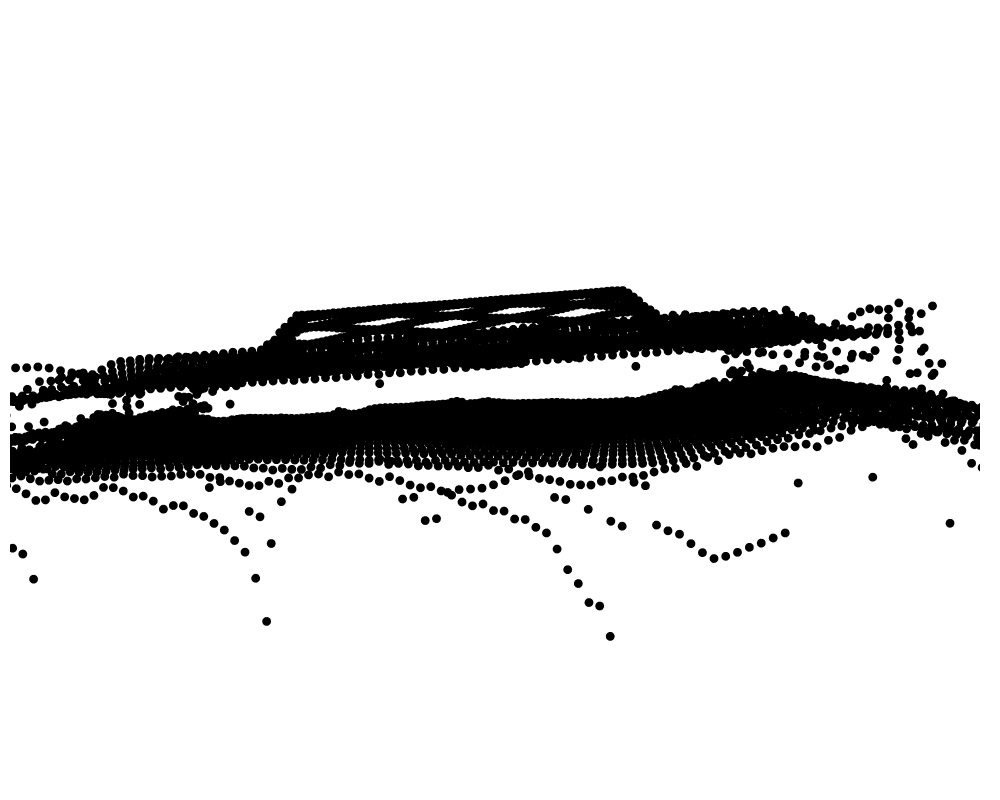}}
    {PSBL\\ w/$\,$norm}
&
\subf{\includegraphics[width=0.18\textwidth]{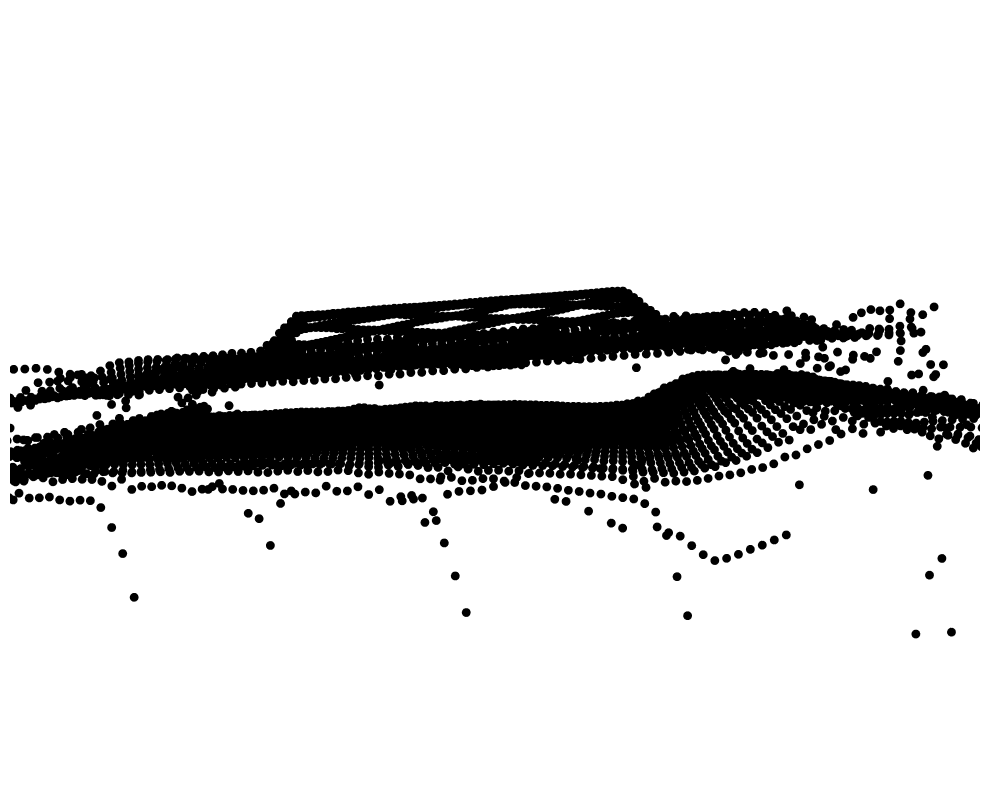}}
     {PSBL\\ w/o norm}
\end{tabular}
\caption{Examples for the reconstruction performance of models trained on different losses on synthetic point clouds if significant parts of key features are occluded. The first column shows the ground truth point clouds from the same view. The key features are marked with a red circle. The visual performance on the synthetic images is very similar for all losses. The model trained with MSE loss shows small artifacts, e.g., on the lower left side of the upper example. The model with the normalized loss function lacks behind its counterpart without normalization regarding the ramp-like edge in the bottom example, which is flatter compared to the ground truth. The best visual reconstruction is given by the models trained with the MAE and the PSBL loss function without normalization.}
\label{syn_images}
\end{figure}

In this section, we will compare the performance of the models trained on MSE, MAE, and the PSBL loss. $\mathcal{L}_{psbl}$ is evaluated with and without the per-layer normalization. The parameters $\alpha$ and $\beta$ are determined by random search in the range $[0.1,1000]$, optimizing for the best PSNR on the evaluation set. With and without normalization, the best loss function is given by:
\begin{equation}
\mathcal{L}_{psbl} = \mathcal{L}_{pixel} + 0.715 \mathcal{L}_{perceptual} + 6.21 \mathcal{L}_{style}
\end{equation}
The visual comparison of the models' outputs in Figure \ref{syn_images} shows that the reconstruction on synthetic point clouds is sufficient for all loss functions.
However, the performance of the models trained solely with MSE loss is worse than the others, followed by the performance of the model with PSBL loss and layer-wise loss normalization. MAE and PSBL loss without normalization generate very similar results.
The lack in performance by the models with MSE loss is also captured by the per-pixel metrics in Table \ref{tab:syn_metrics}.
\begin{table}[tb!]
\centering
\caption{Comparison of the performance of the models trained with different loss functions evaluated on per-pixel metrics. The table shows the mean and variance using different random seeds. The models trained on MAE and MSE loss generate the best performance on the corresponding metric. The best mean PSNR is achieved with the model trained on the PSBL loss with layer-wise loss-normalization. The best SSIM index is achieved by the model trained with PSBL without layer-wise loss-normalization.}
 \begin{tabular}{||c | c | c | c | c||}
 \hline
  & MSE & MAE & PSBL w/$\,$norm & PSBL w/o norm \\ [2ex] 
 \hline\hline
 MSE & \textbf{7.63e-4$\pm$1.6e-10} & 7.84e-4$\pm$1.9e-10 & 7.68e-4$\pm$2.0e-12 & 7.69e-4$\pm$1.9e-10 \\
 MAE & 3.93e-3$\pm$3.3e-7 & \textbf{3.16e-3$\pm$2.9e-11} & 3.78e-3$\pm$2.1e-7 & 3.2e-3$\pm$7.5e-9 \\ 
 PSNR & 33.11$\pm$2.0e-3 & 32.94$\pm$7.5e-3 & \textbf{33.14$\pm$3.8e-3} & 33.13$\pm$9.8e-3 \\
 SSIM & 0.958$\pm$2.5e-5 & \textbf{0.963$\pm$6.4e-6} & 0.961$\pm$5.8e-6 & \textbf{0.963$\pm$4.1e-7} \\
 [1ex]
 \hline
 \end{tabular}
\label{tab:syn_metrics}
\end{table}
The visually perceived lower reconstruction quality of the model trained with PSBL loss with normalization is only indicated by the structural similarity (SSIM) index \cite{wang_ssim}. The per-pixel metrics lack expressiveness as an evaluation metric. However, these are the most suitable options for preselecting the best hyperparameters and model checkpoints with reasonable effort. We decided to optimize our setup based on the PSNR.
\subsection{Performance on Real Data}
\begin{figure}[tb!]
\centering
\begin{tabular}{c| c c c c}
\subf{\includegraphics[width=0.18\textwidth]{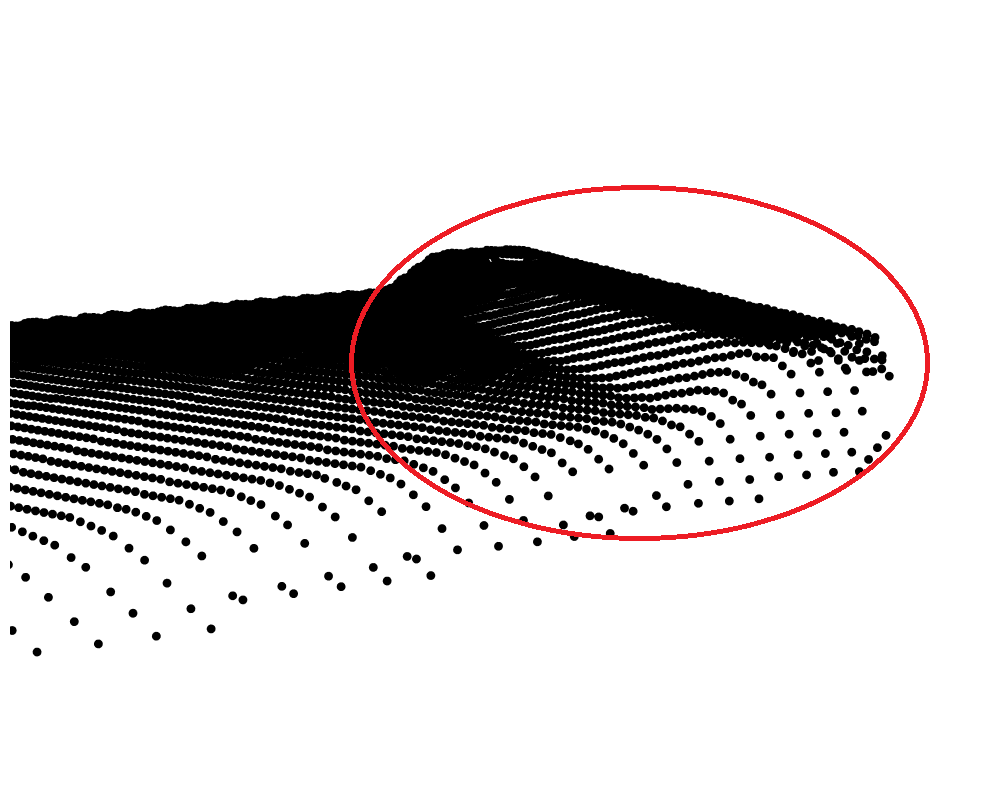}}
     {}
&
\subf{\includegraphics[width=0.18\textwidth]{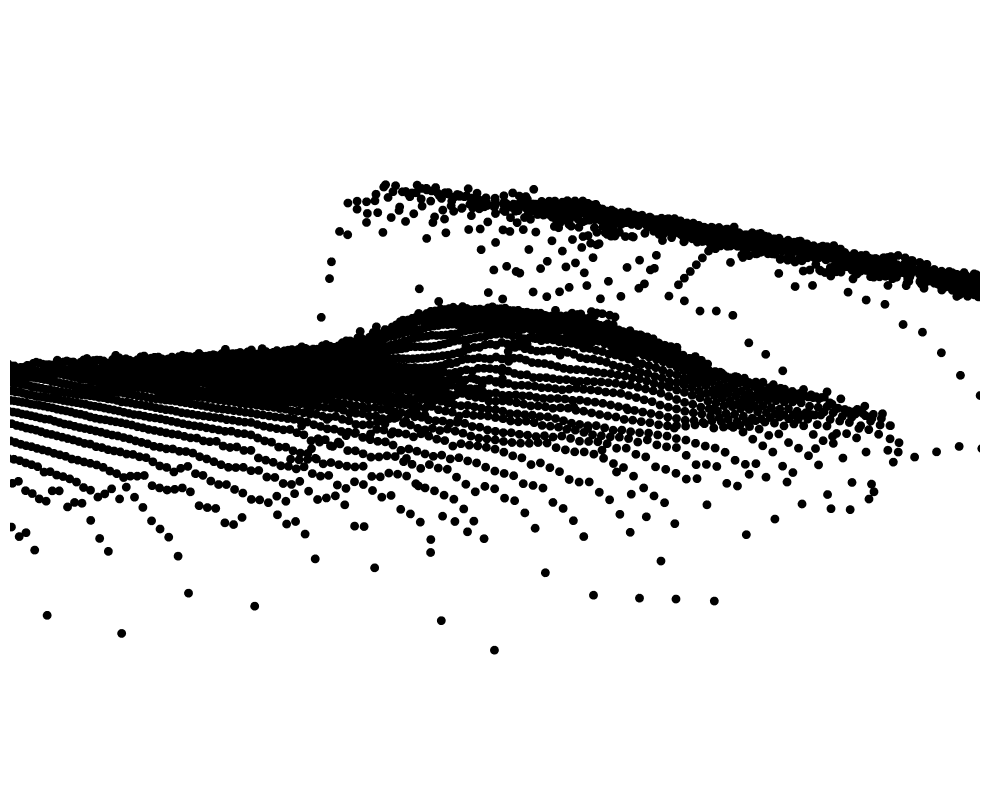}}
     {}
&
\subf{\includegraphics[width=0.18\textwidth]{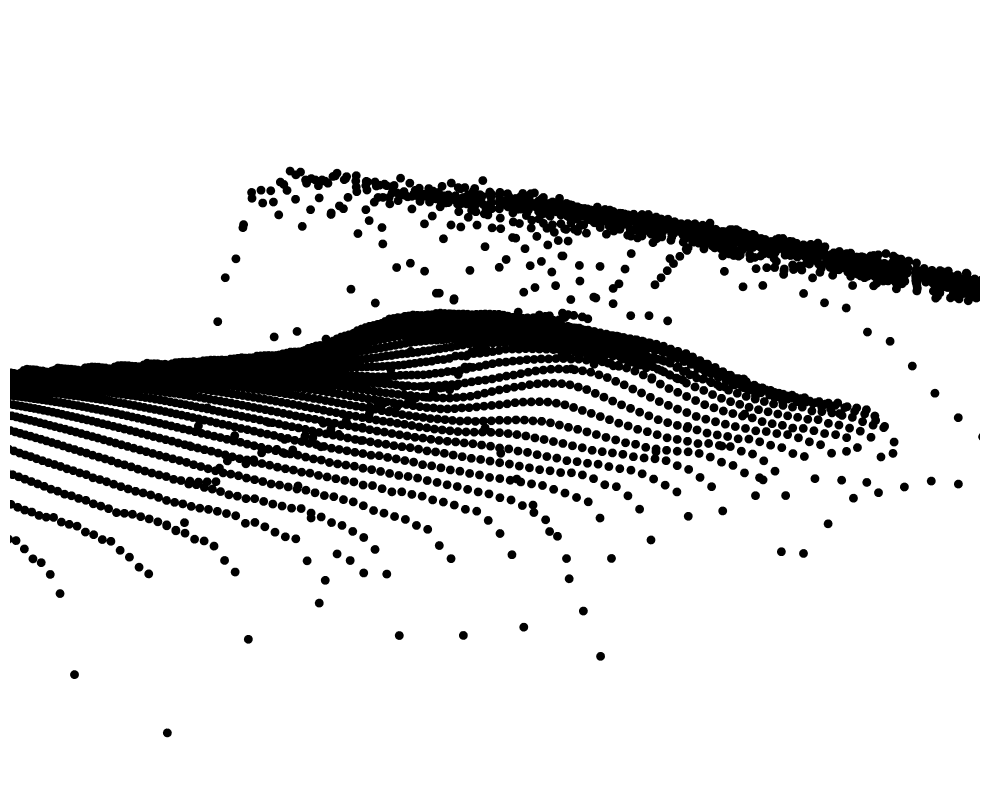}}
     {}
&
\subf{\includegraphics[width=0.18\textwidth]{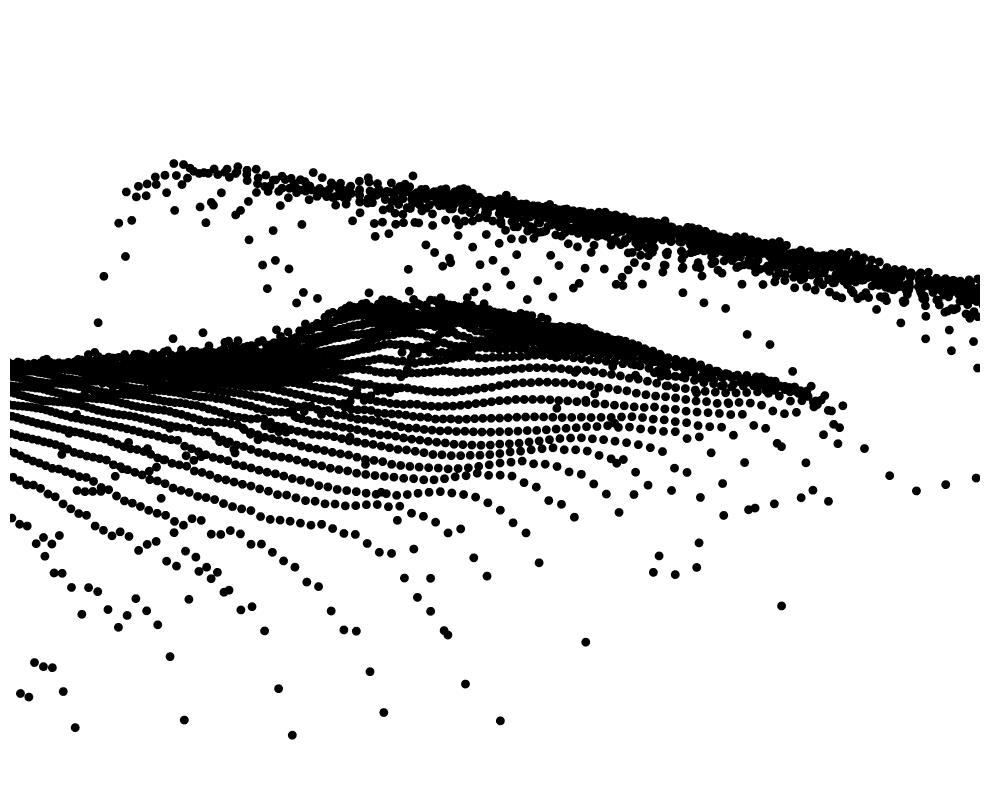}}
     {}
&
\subf{\includegraphics[width=0.18\textwidth]{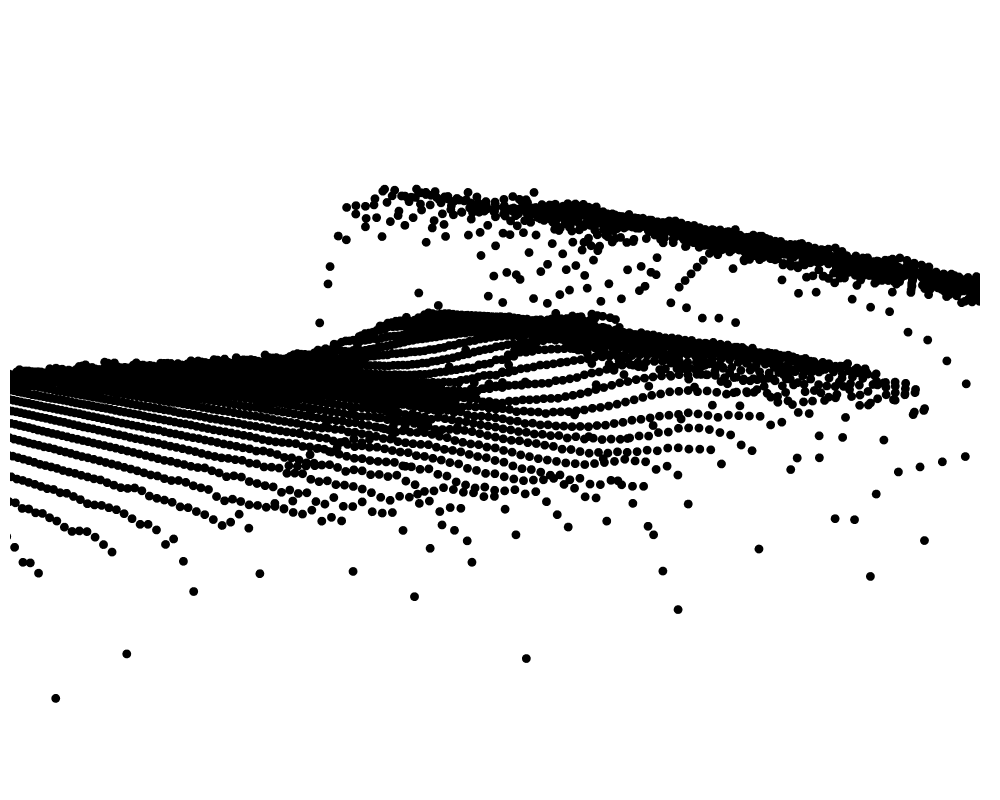}}
     {}
\\
\subf{\includegraphics[width=0.18\textwidth]{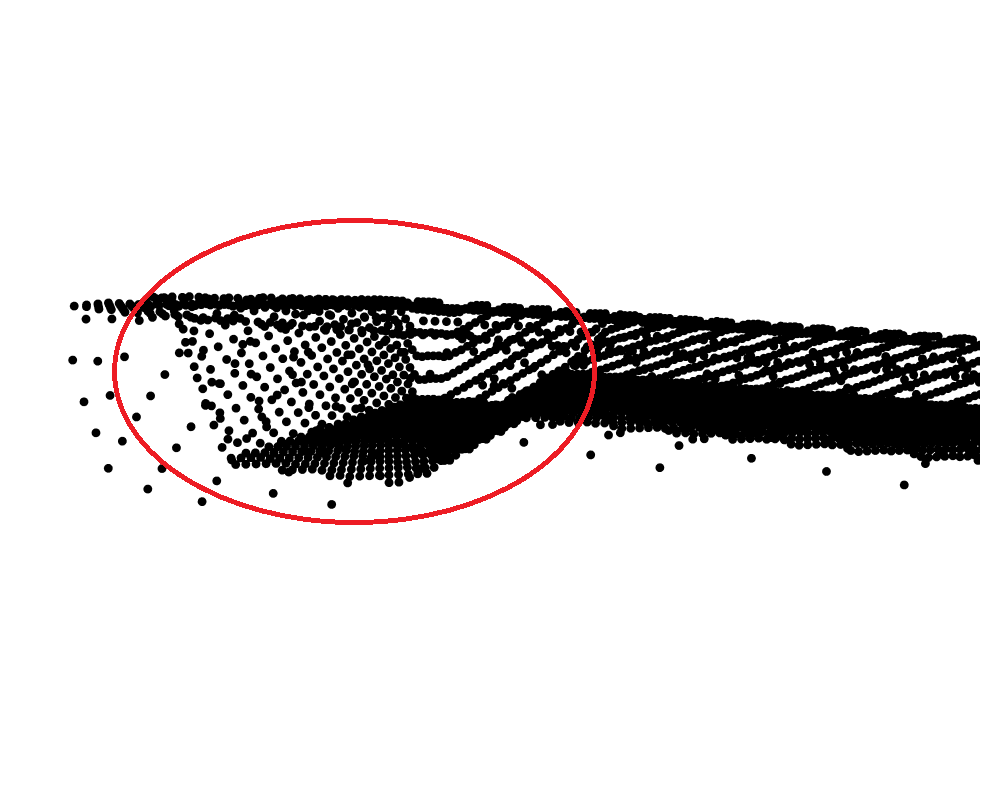}}
     {synthetic \\ reference}
&
\subf{\includegraphics[width=0.18\textwidth]{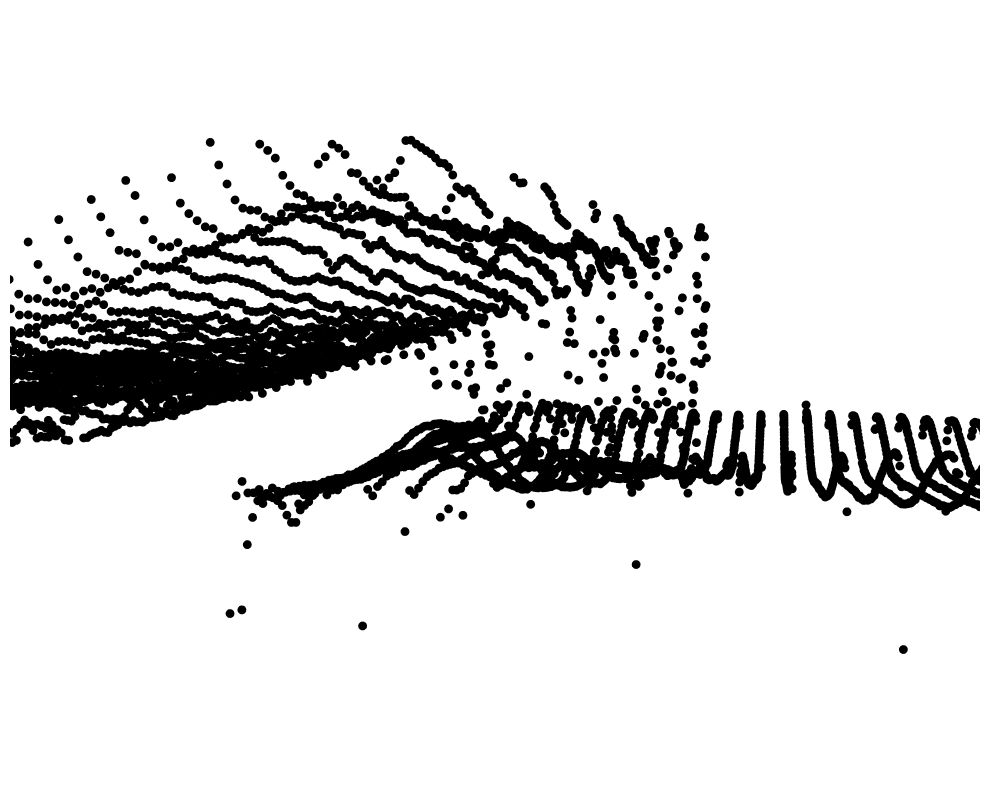}}
     {MSE}
&
\subf{\includegraphics[width=0.18\textwidth]{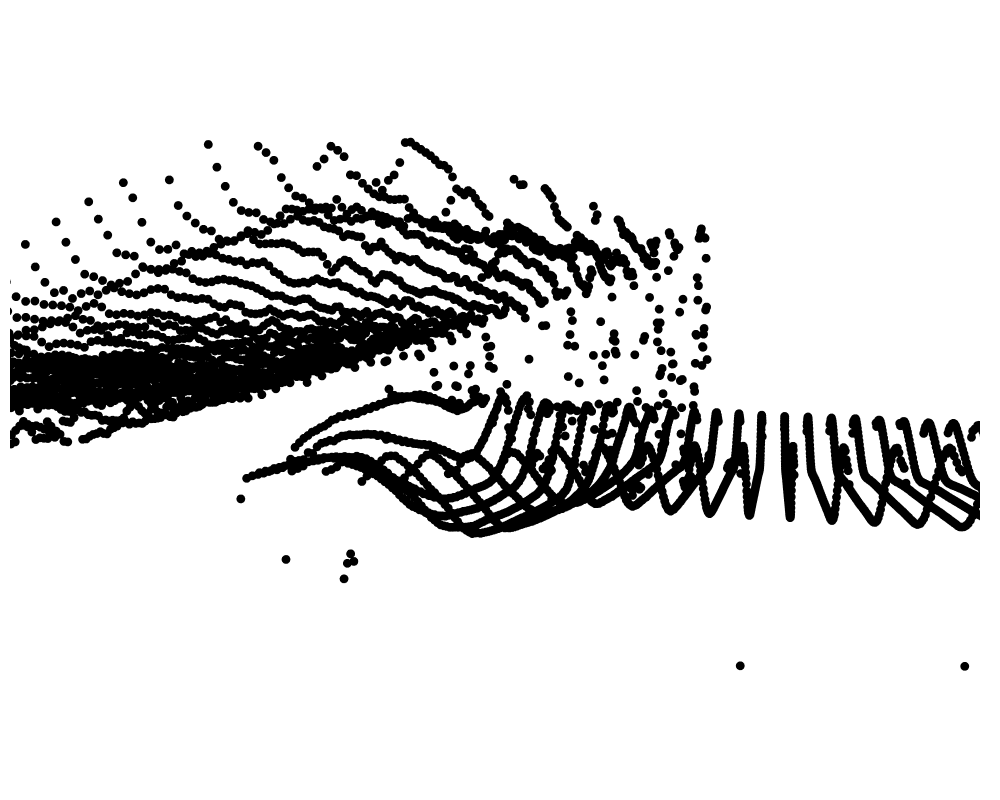}}
     {MAE}
&
\subf{\includegraphics[width=0.18\textwidth]{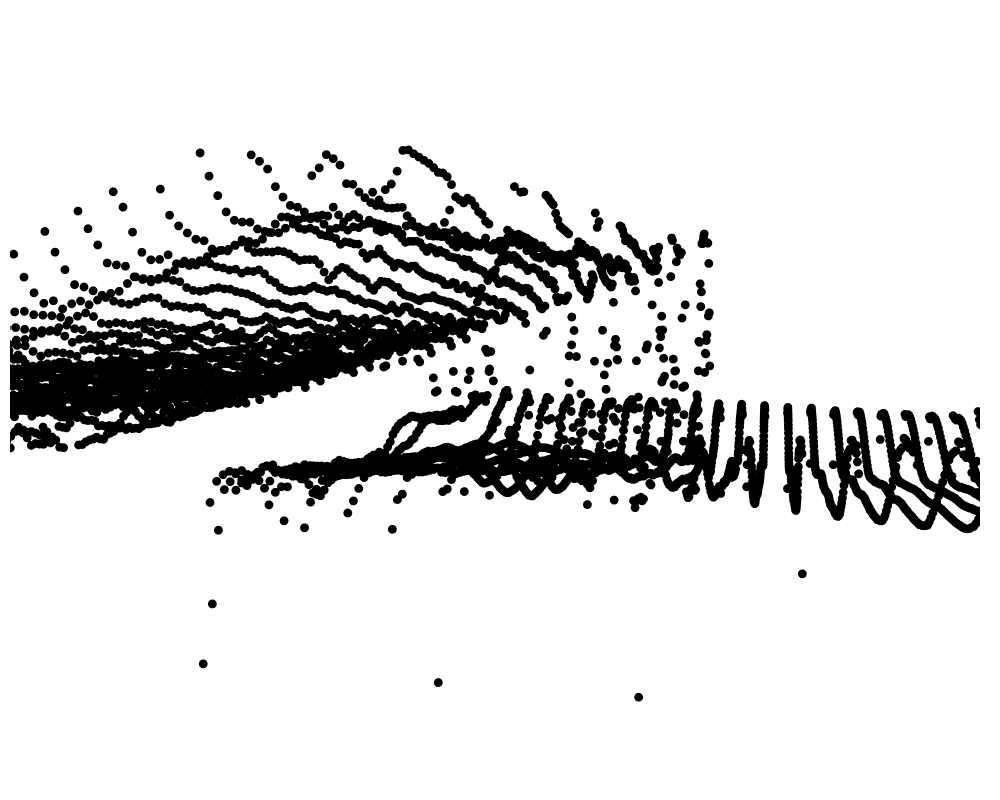}}
    {PSBL\\ w/$\,$norm}
&
\subf{\includegraphics[width=0.18\textwidth]{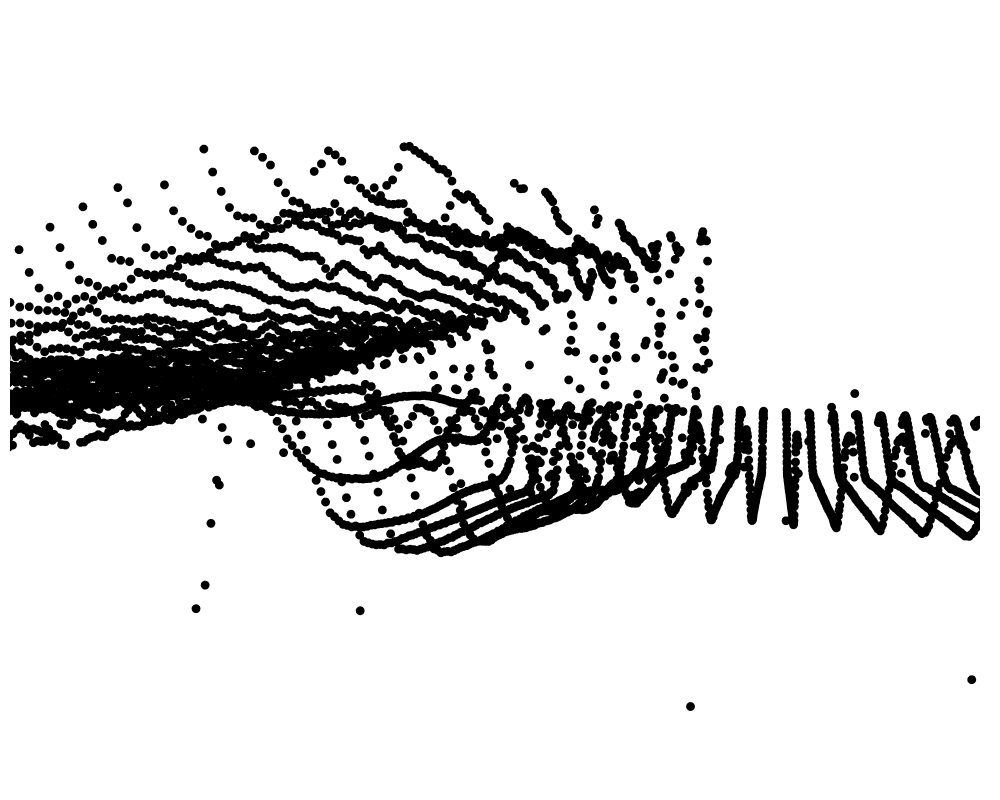}}
     {PSBL\\ w/o norm}
\end{tabular}
\caption{Examples for the reconstruction performance of models trained on different losses on real point clouds if significant parts of key features are occluded. The first column shows synthetic reference point clouds of the lower object. The key features are marked with a red circle. The reference point clouds are generated by manually positioning a synthetic object. The model trained with MSE loss performs worth in reconstructing the marked key features in both shown examples. Especially in the lower example, the quality of the reconstruction of the model with PSBL loss without normalization outperforms the one with normalization.}
\label{real_images}
\end{figure}

For the evaluation on the real data, no ground truth point clouds of the complete lower objects in the proper position exist. For this reason, the per-pixel metrics can not be evaluated.
In Figure \ref{real_images} we show visual examples of the reconstruction performance of the different models on real point clouds.
The reconstruction errors and performance differences are more significant than those on the synthetic data.\newline
Although the models trained with MAE loss and with PSBL loss without normalization have a very similar SSIM index on synthetic images, the latter significantly outperforms the other models in the visual reconstruction quality of the occluded features on real examples with severe occlusions.
This difference is expected to impact the capability of the perception system to match the 3D objects. This directly transfers to the success rate for the calculation of the grasping points and, finally, the grasping process.
The performance on most simpler examples with less severe occlusion is good for all models.
\subsection{Performance of the Pipeline on the Real Process}
\begin{figure}[tb!]
\centering
\begin{tabular}{c c}
\hline
\subf{\includegraphics[width=0.4\textwidth,height=0.1\textheight]{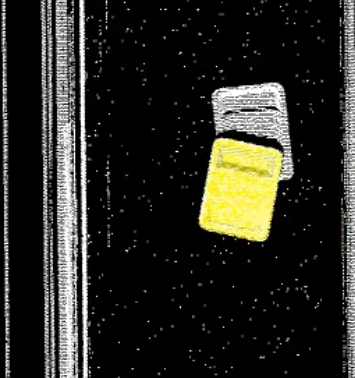}}
     {(a)}
&
\subf{\includegraphics[width=0.4\textwidth,height=0.1\textheight]{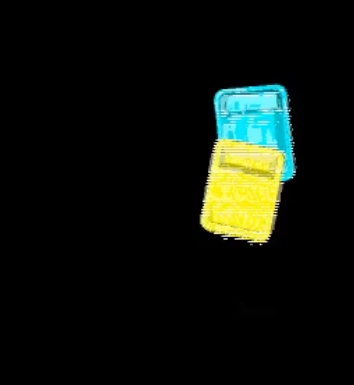}}
     {(b)}
\\
\subf{\includegraphics[width=0.4\textwidth]{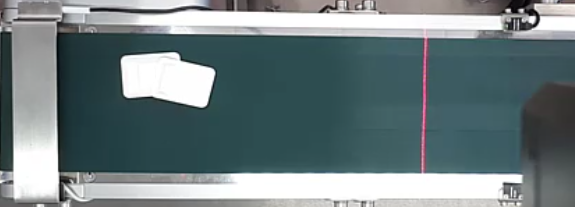}}
     {(c)}
&
\subf{\includegraphics[width=0.4\textwidth]{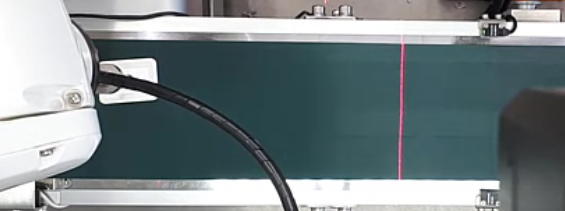}}
     {(d)}
\end{tabular}
\caption{Our solution deployed on the real machine. (a) shows the output of the perception system before the inpainting is executed. In this case, only the yellow 3D shape of the top object is matched. The process detects this and our inpainting solution is triggered and executed. Our pipeline returns the inpainted point cloud (b), where both 3D shapes can be matched. (c) shows the overlapping objects on the conveyor belt. In (d), the grasping process of the lower object is illustrated after matching both objects successfully.}\label{real_process}
\end{figure}

We deployed our solution consisting of the processing pipeline and the models for segmentation and inpainting to the real process. This enabled the grasping application to also pick objects in configurations such as shown in Figure \ref{real_process} that would previously be discarded.
The deployed version is tested on $100$ grasping trials with varying occlusions. The occlusions are manually created by placing one object upon the other in random configurations. In $76\%$ of the examples both objects could be grasped successfully. This shows that we can reduce the amount of discarded objects by $76\%$, thereby drastically increasing the efficiency of the process. The errors in the remaining $24\%$ are distributed across all components, mainly the segmentation model, the inpainting model, and the matching algorithm. We observed that the matching algorithm produces most errors if the surfaces of the reconstructed object and the top object are very close. This happens if the upper object does not stand out at an angle.
The real machine was not available for comprehensive testing. The solution deployed on the real machine uses an inpainting model with $\alpha = 24$ and $\beta=240$ and is trained on $16\,800$ samples. 
Detailed tracking of the exact cause of the errors has not been performed because the generation of the required ground truth in the real process was not possible.

\section{Conclusion}
Conventional real-world grasping applications often face shortcomings if very strict requirements are not met. With the discussed solution, we could solve one of these shortcomings by inpainting missing information caused by occlusions in single-view point clouds. We developed an auxiliary autoencoder model and a sophisticated data processing pipeline. By reducing the task complexity, we could train our model solely on synthetic data. The deployed solution achieved to pick $76\%$ of otherwise discarded objects in a real-world process.\newline
We see many opportunities for facilitating well-established applications in manufacturing environments using ML to handle cases they currently fall short. At the same time, one could still rely on the tested performance of the established methods in those cases where they already work reliably.

\subsubsection{Acknowledgements} We would like to thank Robert Schmeisser, Harald Funk, and all associated colleagues for making the project possible. We are especially grateful to them for setting up the simulation and helping deploy the solution on a real machine.

\end{document}